\documentclass[10pt,journal]{IEEEtran}
\IEEEoverridecommandlockouts 
\usepackage{blkarray}                                      
\usepackage{algpseudocode}                                 
\usepackage{algorithm}
\usepackage{graphicx}                                      
\usepackage{amsmath}
\usepackage{amssymb}
\usepackage{amsfonts}
\usepackage{amsthm}
\usepackage[mathcal]{eucal}
\usepackage{mathrsfs}
\usepackage{booktabs}
\usepackage{enumerate}
\usepackage{multirow}
\usepackage[subrefformat=parens,farskip=0pt,justification=centering]{subfig}
\usepackage{color}
\usepackage{cite}                                          
\usepackage{comment}                                       
\usepackage{soul}                                          
\soulregister\cite7
\soulregister\ref7
\soulregister\pageref7
\usepackage{etoolbox}                                      
\usepackage{url}
\usepackage{nth}                                           
\usepackage{bm}                                            
\usepackage{courier}
\usepackage{balance}
\usepackage{threeparttable}
\usepackage{xcolor,colortbl}
\usepackage{footnote}
\usepackage{listings}
\usepackage{setspace}                                      

\usepackage{verbatim}
\usepackage[bookmarks=false]{hyperref}
\hypersetup{
    colorlinks = true,
    citecolor  = blue,
    linkcolor  = blue,
    urlcolor   = blue,
}
\usepackage{tikz}
\usetikzlibrary{patterns,snakes}
\usetikzlibrary{positioning,calc,fit,decorations.pathmorphing,shapes.geometric, shapes.gates.logic.US, calc}
\usetikzlibrary{arrows,arrows.meta,decorations.markings,shapes,shapes.arrows}
\usetikzlibrary{decorations,decorations.pathreplacing}
\usetikzlibrary{backgrounds}
\usepackage{filecontents}                                  
\usepackage{pgfplots}
\usepackage{pgfplotstable}
\usepackage{scalefnt}
\pgfplotsset{compat=newest}
\usepackage{caption}
\usepackage{cleveref}
\Crefformat{figure}{Fig.~#2#1#3}                           
\Crefname{subfigure}{Fig.}{Figs.}
\Crefname{figure}{Fig.}{Figs.}
\Crefformat{table}{TABLE~#2#1#3}                           
\usepackage[figuresright]{rotating}

\definecolor{CUHKorange}{RGB}{244,106,18} 
\definecolor{CUHKblue}{RGB}{0,111,190}    
\definecolor{CUHKgreen}{RGB}{0,127,128}   
\definecolor{CUHKred}{RGB}{228,46,36}     
\definecolor{CUHKyellow}{RGB}{198,148,34} 
\definecolor{CUHKdark}{RGB}{114,44,114}   
\definecolor{CUHKmiddle}{RGB}{144,44,144} 
\definecolor{CUHKlight}{RGB}{167,44,167} 
\definecolor{CUHKpurple}{RGB}{117,15,109}
\definecolor{CUHKgold}{RGB}{221,163,0}
\definecolor{CUHKribbon}{RGB}{244,223,176}
\definecolor{CUHKblack}{RGB}{34,24,21}

\usepackage{tcolorbox}
\tcbuselibrary{skins,breakable}
    {\endtcolorbox}

\usepackage[top=0.75in,bottom=0.80in,left=0.58in,right=0.58in]{geometry}
\crefname{mytheorem}{Theorem}{Theorems}
\crefname{mylemma}{Lemma}{Lemmas}
\crefname{myclaim}{Claim}{Claims}
\crefname{myproperty}{Property}{Properties}
\crefname{mycorollary}{Corollary}{Corollaries}

\algrenewcommand\textproc{\texttt}

\makeatletter
\let\OldStatex\Statex
\renewcommand{\Statex}[1][3]{%
  \setlength\@tempdima{\algorithmicindent}%
  \OldStatex\hskip\dimexpr#1\@tempdima\relax
}
\makeatother

\RequirePackage[normalem]{ulem} 
\RequirePackage{color}\definecolor{RED}{rgb}{1,0,0}\definecolor{BLUE}{rgb}{0,0,1} 

\graphicspath{{./figs/}{../}}
\begin{document}

\title{Efficient and Versatile Visual Knowledge Integration into Pre-Trained Language Models }
\author{Xinyun Zhang, \quad
    Haochen Tan,  \quad
    Han Wu,       \quad
    Bei Yu
    \thanks{Xinyun Zhang and Bei Yu are with the Department of Computer Science and Engineering, The Chinese University of Hong Kong, NT, Hong Kong SAR.}
    \thanks{Haochen Tan, Han Wu are with the Department of Computer Science, City University of Hong Kong, Kowloon, Hong Kong SAR.}
}

\maketitle
\pagestyle{plain}
\begin{abstract}
    Humans learn language via multi-modal knowledge. However, due to the text-only pre-training scheme, most existing pre-trained language models (PLMs) are hindered from incorporating multi-modal information.
    To integrate visual knowledge into PLMs, existing methods require updating all the original parameters of PLMs for knowledge fusion, and 
 they only incorporate either the text or image encoder of vision-language models (VLMs) to encode visual information.
    In this paper, we propose a new plug-and-play module, X-adapter, to flexibly leverage the aligned visual and textual knowledge learned in pre-trained VLMs and efficiently integrate them into PLMs.
    Specifically, we insert X-adapters into PLMs, and only the added parameters are updated during adaptation.
    To fully exploit the potential of VLMs, X-adapters consist of two sub-modules, V-expert and T-expert, to fuse VLMs' image and text representations, respectively.
    We can activate different sub-modules depending on the downstream tasks.
    Experimental results show that our method can significantly improve the performance of PLM baselines on object-color reasoning and natural language understanding (NLU) tasks.
\end{abstract}

\section{Introduction}
Pre-trained language models have achieved great success on many NLP tasks \cite{BERT, wang2018glue, gpt3, radford2019language}. By predicting missing tokens based on the context, masked language modelling (MLM) \cite{BERT} explores the self-supervision potential in massive unlabeled text data and spawns a series of powerful pre-trained models, such as BERT \cite{BERT}, RoBERTa \cite{liu2019roberta}, ALBERT \cite{lan2019albert}, and ELECTRA~\cite{clark2020electra}.


Albeit significant progress has been made, merely learning from the textual context prevents the language models from acquiring commonsense knowledge seldomly seen in the text corpus, e.g., the visual appearance of different objects \cite{jin2022leveraging}. The lack of multi-modal knowledge can lead to false predictions regarding reasoning tasks related to visual understanding \cite{wang2022visually,logan-etal-2019-baracks}. For example, BERT will answer "red" if we ask it the color of a banana. 

\begin{figure}[tb!] 
    \centering
    \includegraphics[width=\linewidth]{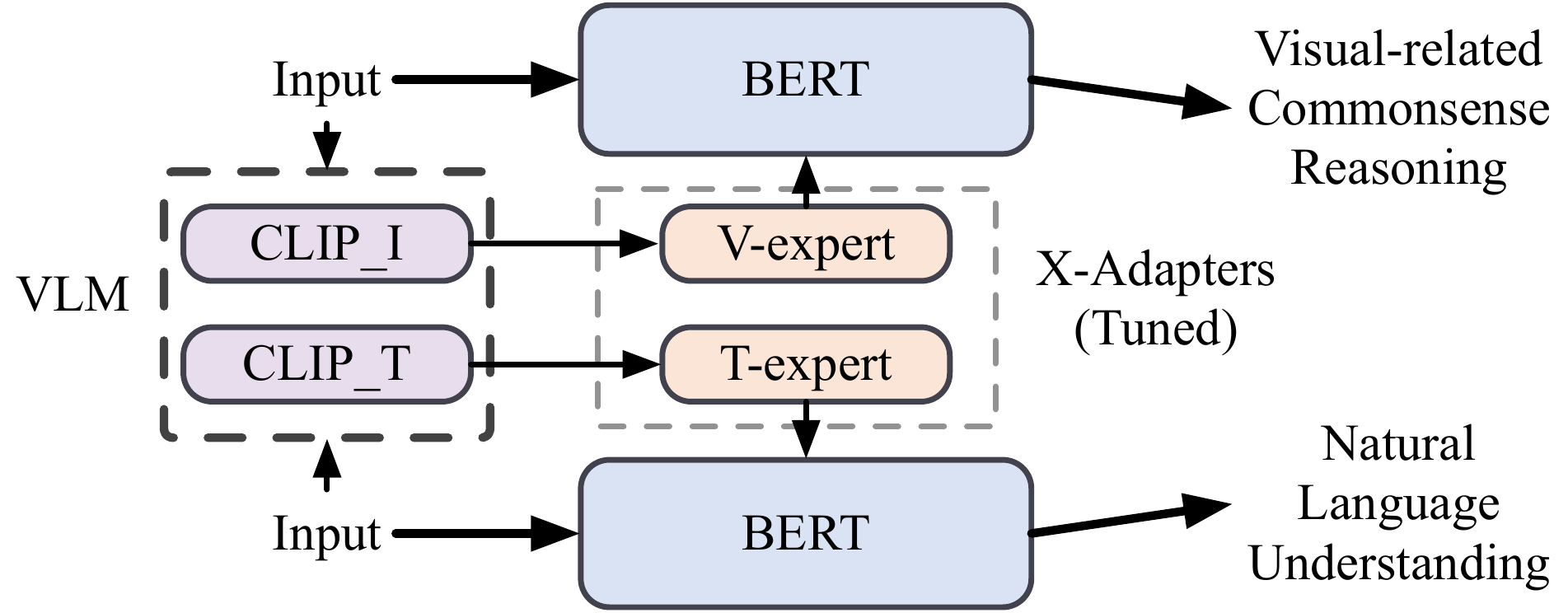}
    \caption{The main idea of X-adapters. For different downstream tasks we activate different sub-modules in X-adapters to fully exploit the VLMs. During adaptation, only X-adapters' parameters are updated.} 
    \label{fig:intro}
\end{figure}

To mitigate this issue, a common practice is to import the multi-modal knowledge learned in VLMs into PLMs. A widely adopted choice for VLMs is the dual-encoder models, e.g., CLIP \cite{clip}, consisting of two well-aligned encoders for image and text, respectively. Previous works \cite{vokenization, vidlankd, xdbert, jin2022leveraging} distil visual knowledge from VLMs' text encoder to masked language models, during either pre-training or intermediate pre-training \cite{jin2022leveraging, xdbert}. Meanwhile, VaLM \cite{wang2022visually} incorporates features encoded by CLIP's \cite{clip} image encoder into pre-training causal language models. However, these methods all require updating \textbf{all} the original LM parameters and even the CLIP text encoder \cite{xdbert}, resulting in a high memory footprint, especially for large language models. Moreover, they merely keep an eye on \textbf{one side} of VLMs (image or text encoder), which does not fully exploit VLMs' potential and thus leads to improvements only on either visual-related commonsense reasoning \cite{wang2022visually, jin2022leveraging} or NLU tasks \cite{xdbert}. These limitations significantly hinder a broader application of visual knowledge fusion into PLMs. 


To bridge these gaps, we propose a novel framework for efficient and versatile visual knowledge integration into pre-trained masked language models, as shown in \Cref{fig:intro}. Specifically, we propose a new plug-and-play module, dubbed X-adapter, to flexibly fuse the features from the text and image encoders of pre-trained VLMs \cite{clip}. Two sub-modules of X-adapters, V-expert and T-expert, account for injecting the image and text representations from VLMs, respectively. Depending on the downstream tasks, we can activate different sub-modules to inject appropriate visual knowledge. Given a pre-trained masked language model, we first insert several X-adapters into the transformer encoders. Then, only the parameters of the X-adapters are updated during adaptation, which significantly reduces the memory footprint, as shown in \Cref{tab:param-and-memory}.

We conduct extensive experiments to validate the effectiveness of our proposed method. To verify models' capabilities in reasoning visual commonsense concepts, we conduct zero-shot reasoning experiments on object colors, similar to \cite{wang2022visually}. Experimental results demonstrate that our method can outperform the baseline PLMs by approximately 30\%, activating the V-expert in X-adapters. In addition, we also conduct experiments on NLU tasks \cite{wang2018glue}, and our method can surpass the baselines with a substantial margin by activating the T-expert in X-adapters.

\begin{figure*}
    \centering
    \subfloat[]{ \includegraphics[height=6.8cm]{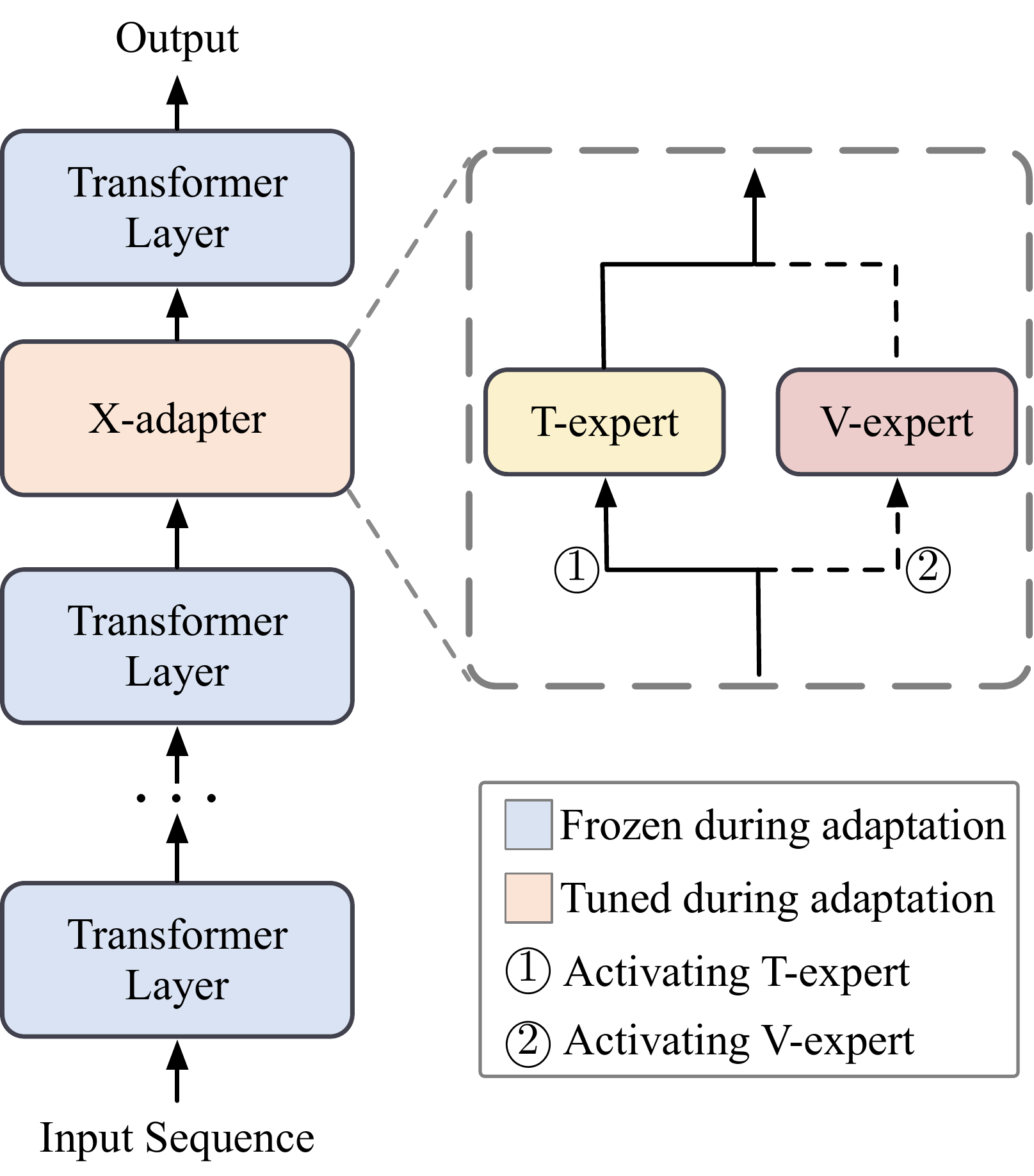} \label{fig:main-arch} } \hspace{0.08cm}
    \subfloat[]{ \includegraphics[height=6.8cm]{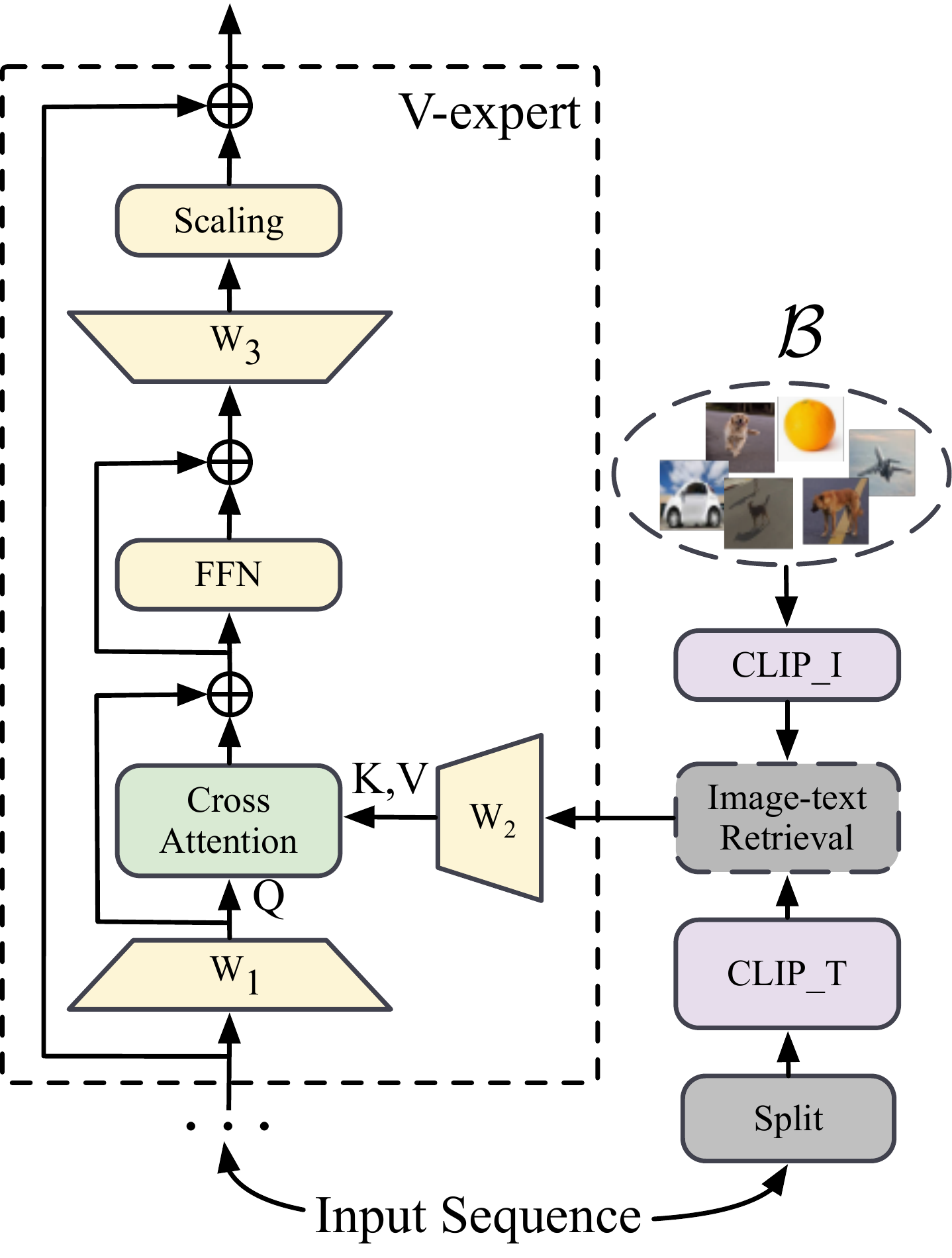}          \label{fig:v-expert} }  \hspace{0.08cm}
    \subfloat[]{ \includegraphics[height=6.8cm]{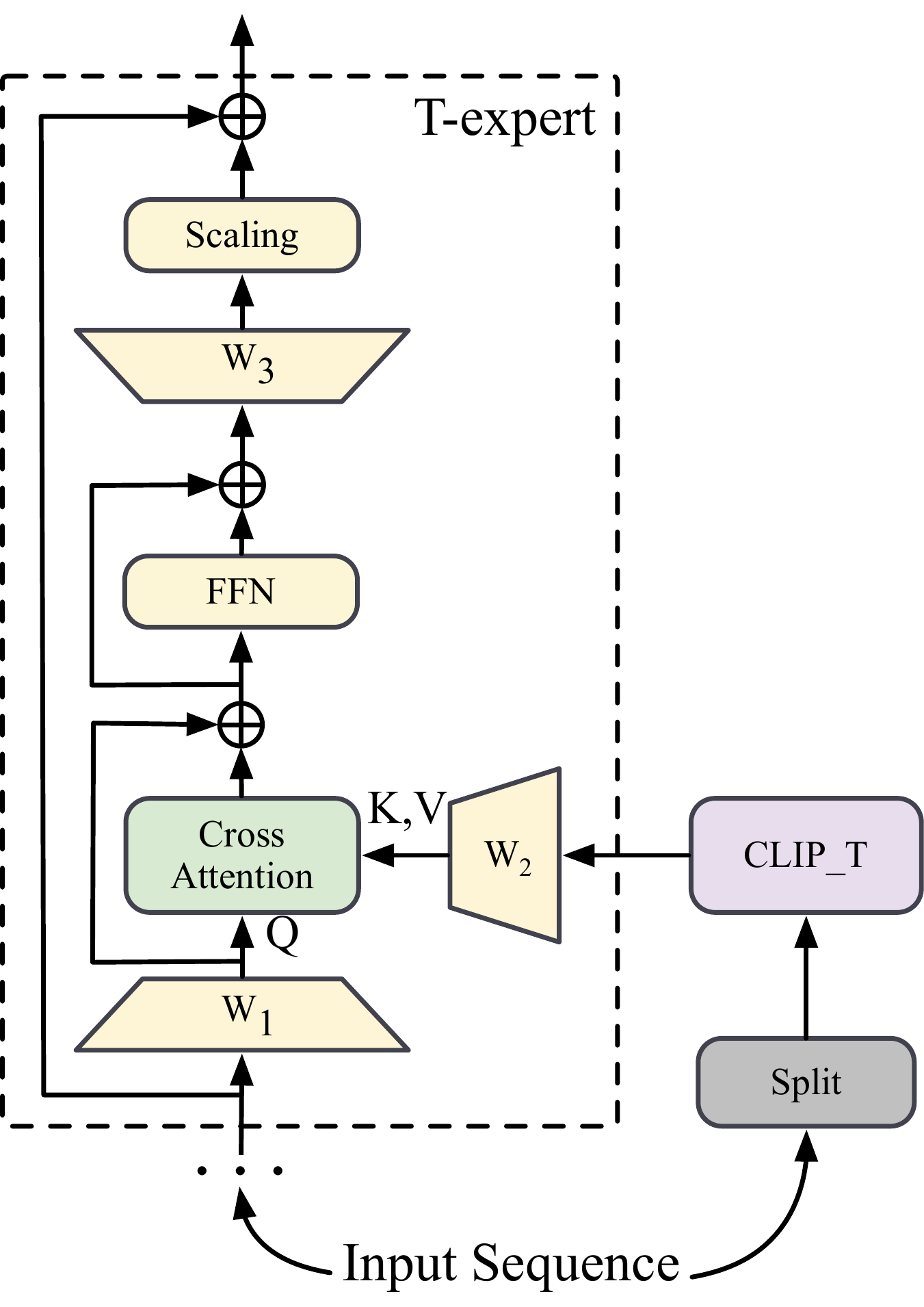}          \label{fig:t-expert} }
        \caption{(a): The main architecture of our proposed method; (b): The detailed architecture of V-expert; (c): The detailed architecture of T-expert.}
    \label{fig: mlm-ratio}
\end{figure*}

We summarize our contributions as follows:
\begin{itemize}
\item We shed light on the problem of visual knowledge injection into PLMs and propose a new module, X-adapter, to fully exploit VLMs' image and text representations. 

\item By only updating the inserted parameters, our method can efficiently adapt the visual knowledge into the pre-trained language models with much less memory footprint.

\item Extensive results demonstrate that our method can outperform the baseline models on color reasoning tasks by a margin of approximately 30\% while achieving notable improvements on NLU tasks simultaneously.
\end{itemize}




\begin{table}[!tbp]
    \centering
    {
        \begin{tabular}{c|c}
            \toprule
             Method  & Memory Usage (GB)  \\
            \midrule
             Co-training \cite{xdbert} & 80.0 \\
             Distillation \cite{vokenization} & 57.3 \\
             X-adapter-L2 & 30.9 \\
            \bottomrule
        \end{tabular}
    }
   \caption{The comparison of GPU consume for visual-knowledge injection methods. We insert two X-adapter layers and activate the T-expert. We fix the VLM and input length to be the same for all three methods.}
   \label{tab:param-and-memory}
\end{table}

\section{Related Work}
\subsection{Vision-language Models}
Vision-language models map the text and image features into a unified representation space and pave the way for fusing visual information into language models. Early attempts \cite{chen2020uniter,li2019visualbert,su2019vl,li2020oscar,tan2019lxmert} train a unified cross-modal encoder to close the gap between the vision and language feature spaces. They treat the image patches and the language tokens uniformly and learn their interaction through a stack of transformer encoders. Normally, the training objective is masked language/image modeling \cite{BERT}, which randomly masks a portion of input and guides the model to predict the missing parts based on the context. These methods achieve significant progress on a various of vision-language downstream tasks, such as visual question answering (VQA). Recently, contrastive learning-based methods \cite{clip,jia2021scaling} have significantly pushed the boundaries of vision-language representation learning. As a widely used representative, Contrastive language image pre-training (CLIP) \cite{clip} trains a text and an image encoder with an in-batch contrastive loss to align the text and image representations. Specifically, they first collect about 400M image-text pairs from the web as the training data. Then, for each iteration, they input a large number of image-text pairs as a batch. The text and image encoder encode the text and image input, respectively. For each image/text feature, only the corresponding text/image feature will be considered as the positive sample while the others will be considered as the negative samples, and the contrastive loss will pull together the positive samples while pushing apart the negative samples. Benefiting from the vast amount of image-text pairs, CLIP learns robust and well-generalizable vision and language representations, which benefit many vision or vision-language tasks, such as zero-shot image recognition \cite{clip} and text-to-image generation \cite{ramesh2022hierarchical}. In this paper, we also leverage CLIP as the pre-trained VLM from which we transfer the visual knowledge to the PLMs.

\subsection{Visually-enhanced Language Models}
Many efforts have been made to incorporate visual information into language models as compensation for the lack of commonsense knowledge during text-only pre-training \cite{vokenization, vidlankd, jin2022leveraging,xdbert,wang2022visually,iace}.
One line of work focuses on injecting visual knowledge during pre-training language models.
Vokenization \cite{vokenization} and VidLanKd \cite{vidlankd} train a vision-language model and distil knowledge from its text encoder. VaLM \cite{wang2022visually} first retrieves the relevant visual features from CLIP's visual encoder \cite{radford2021learning} and then appends them to the language tokens as the input of the transformer layers. Although significant progress has been made, these works all requiring a new language model \textbf{from scratch} to fuse the visual knowledge into the language models, and it is infeasible for them to enhance a well pre-trained language model.
Another line of work investigates how to effectively transfer visual knowledge to PLMs. \cite{jin2022leveraging} investigates visual knowledge fusion through intermediate pre-training, which re-trains the PLMs on a relatively smaller amount of vision-language data compared with full pre-training. Under this setting, they observes that distillation from VLM's text encoder improves the performance on visual-related commonsense reasoning tasks but also leads to degradation on NLU tasks. Similarly, \cite{xdbert} proposes a co-training scheme, named XDBert, which tunes both the LM and the VLM to transfer the knowledge from the CLIP textual encoder on a smaller amount of image-text pairs, achieving improvements on NLU tasks. Although the intermediate pre-training setting they consider is much more effective then full pre-training, both methods still require tuning all the parameters in the PLMs, which can cause large computational cost, especially for the large PLMs. 
Following this line of work, we propose a more effective and efficient way to fully adapt text and image representations in VLMs, which only requires tuning much fewer parameters compared with tuning the whole PLM and leads to improvements in both visual-related commonsense reasoning and NLU tasks. 



\subsection{Efficient Knowledge Transfer}
Previous works have explored the efficient knowledge transfer in different tasks and models \cite{houlsby2019parameter, hu2021lora, he2021towards, wang2020k}. Freezing the pre-trained models, \cite{he2021towards} proposes to insert adapters into the frozen models, which enables parameter- and computation-efficient knowledge transfer from the pre-trained models to the downstream tasks. Similarly, \cite{wang2020k} proposes to plug adapters in pre-trained language models and update the corresponding parameters to inject some factual or linguistic knowledge. In this paper, we propose a new adapter architecture, X-adapter, to fully exploit the competence of pre-trained vision-language models and efficiently transfer the multi-modal knowledge into the pre-trained language models.

\section{Method}
The main framework of our proposed method is shown in \Cref{fig:main-arch}. In the following sections, we will detail X-adapter's architectures and the learning procedure, including the adaptation training, zero-shot reasoning and fine-tuning for downstream tasks.


\subsection{X-adapters}
As shown in \Cref{fig:main-arch}, an X-adapter consists of two sub-modules, V-expert and T-expert, integrating the CLIP's visual and textual features, respectively. Both the V-expert and the T-expert take a similar architecture as shown in \Cref{fig:v-expert} and \Cref{fig:t-expert}. Given a query vector $\bm{x} \in \mathbb{R}^{1 \times d}$ in PLM where $d$ is the model dimension of the transformer, we first use a matrix $ \bm{W}_1 \in \mathbb{R}^{d \times r}$  to project the feature to a lower dimension $r$. Then, we use a multi-head cross-attention module to fuse the visual features $\bm{V} \in \mathbb{R}^{N \times d_c}$ with the query vector, where $d_c$ is the hidden dimension of CLIP, and $N$ is the number of visual features to be injected, depending on the sub-module. The fusion process can be formulated as:
\begin{equation}
\begin{aligned}
    \bm{u} &= \mathrm{MHA}(\bm{x}\bm{W}_1,\bm{V}\bm{W}_2,\bm{V}\bm{W}_2) \\ &=\mathrm{Concat}(\mathrm{head}_1, \cdots, \mathrm{head}_n)\bm{W}_o, \\
\end{aligned}
\end{equation}
where
\begin{equation}
    \mathrm{head}_i = \mathrm{Attn}((\bm{x}\bm{W}_1)\bm{W}_q^{i}, \bm{V}\bm{W}_2\bm{W}_k^{i}, \bm{V}\bm{W}_2\bm{W}_v^{i}),
\end{equation}
$\bm{W}_o \in \mathbb{R}^{r \times r}$, $\bm{W}_2 \in \mathbb{R}^{d_c \times r}$, $n$ is the number of heads and $\bm{W}_q^{i}, \bm{W}_k^{i}, \bm{W}_v^{i} \in \mathbb{R}^{r \times \frac{r}{n}}$ are the projection matrices for the query, key and value in the $i$-th head, respectively.
Then, a shortcut is connected to the feature before the attention, denoted by
\begin{equation}
\tilde{\bm{u}} = \mathrm{LN}(\bm{u} +\bm{x}\bm{W}_1),
\end{equation}
where $\mathrm{LN(\cdot)}$ is layer normalization \cite{ba2016layer}.
Further, we encode the fused feature by a feed-forward network with shortcut:
\begin{equation}
    \bm{m} = \mathrm{LN}(\tilde{\bm{u}} + \mathrm{FFN}(\tilde{\bm{u}})).
\end{equation}
Finally, a projection matrix $\bm{W}_3 \in \mathbb{R}^{r \times d}$ maps the feature back to the transformer model dimension. With a learnable scale parameter $s$ and a shortcut to the input feature, the final output of the X-adapter is:
\begin{equation}
\bm{x}_{out} = \mathrm{LN}(s \cdot \bm{m}\bm{W}_3 + \bm{x}).
\end{equation}
The only difference between V-expert and T-expert lies in selecting the visual features $\bm{V}$. Indeed, V-expert handles the representations from the image encoder of CLIP while T-expert handles that from the text encoder. Now we detail the acquisition procedure of the visual features.

\paragraph{V-expert} For V-expert, the target is to find $K$ most relevant images to the given input sequence and then obtain the visual features via CLIP's image encoder ($\mathrm{CLIP\_I}$), as shown in \Cref{fig:v-expert}. Inspired by \cite{wang2022visually}, the main idea is to collect an image bank $ \mathcal{B}$ and use the textual representations from the CLIP's text encoder as the query to retrieve top $K$ nearest neighbors based on the cosine similarity. However, the input text may consist of multiple sentences, each depicting different objects. In this case, directly using the textual representation of the whole input text may not be able to retrieve all the relevant visual objects for each sub-sentence. Therefore, we first split the input text $\bm{t}$ into a set of sub-sentences $ \{\bm{t}_1,\cdots,\bm{t}_l\}$ where $l$ is the number of sentences using NLTK\footnote{\url{https://www.nltk.org/}}. Then, we evenly retrieve the most relevant images to each sentence to construct the visual feature set for the input text, as shown in \Cref{alg:retrieval}. After obtaining the image features, we stack them into the visual features $\bm{V} \in \mathbb{R}^{K \times d_c}$.

\begin{algorithm}
\caption{Image Retrieval Algorithm}\label{alg:retrieval}
\hspace*{\algorithmicindent} \textbf{Input}: input text $\bm{t}$ \\
 \hspace*{\algorithmicindent} \textbf{Output}: a set $\mathcal{V}$ containing $K$ image features 
\begin{algorithmic}[1]
\State $\mathcal{V} \gets \emptyset$;
\State $\{\bm{t}_1,\cdots,\bm{t}_l\} \gets \mathrm{Split}(\bm{t})$;
\State Randomly select $ \mathcal{I} \subseteq \{1, \cdots, l\}$ such that $\left|\mathcal{I} \right|=K\bmod l$;
\For{$i \in \{1, \cdots, l\}$}
\If{$i \in \mathcal{I}$}
    \State $n \gets \lfloor \frac{K}{l} \rfloor+1$;
\Else
    \State $n \gets \lfloor \frac{K}{l} \rfloor$;
\EndIf
\State $\mathcal{U} \gets$ Retrieve $n$ nearest features of $\bm{t}_i$;
\State $\mathcal{V} \gets \mathcal{V} \cup \mathcal{U}$;
\EndFor

\end{algorithmic}
\end{algorithm}

\paragraph{T-expert} As shown in \Cref{fig:t-expert}, we directly leverage the textual representations encoded by CLIP's text encoder ($\mathrm{CLIP\_T}$) as $ \bm{V}$ for T-expert. Since CLIP's maximum sequence length is 77, which is shorter than the input sequence in many downstream tasks, e.g., NLU tasks, we need to split the input text into chunks that fit in the CLIP's input length limit. Specifically, given an input sequence, we first tokenize it using CLIP's tokenizer and then split the input ids into chunks of length 77, denoted as $\bm{t}_1, \cdots, \bm{t}_n$, where $n$ is maximum chunk number. Then, the textual features of the chunks are concatenated and padded to a fixed length $L$ as the visual features $ \bm{V} \in \mathbb{R}^{L \times d_c}$ for the input sentence.

\subsection{Learning Procedure}

\paragraph{Adaptation} We insert several X-adapter layers into the transformer encoder of PLMs, as shown in \Cref{fig:main-arch}. Then, we adopt Masked Language Modeling (MLM) \cite{BERT} as the objective to adapt the inserted visual knowledge with the language knowledge learned in PLMs. Given a text corpus $\mathcal{T}$, we randomly mask each sentence $\bm{t} \in \mathcal{T}$ following the strategy in \cite{BERT}, denoted as $\tilde{\bm{t}}$. The objective can be formulated as:
\begin{equation}
    \mathcal{L}(\mathcal{T})=\frac{1}{\left| \mathcal{T} \right|}\sum_{\bm{t} \in \mathcal{T}}\sum_{\substack{t_i \in \mathcal{M} \\ 
    \left| \mathcal{M}\right|=m \left| \bm{t} \right| }} \mathrm{log}(t_i|\tilde{\bm{t}}),
\end{equation}
where $\mathcal{M}$ is the masked token set, and $m$ is the mask ratio. Unlike common practice that sets $m$ to a fixed ratio 15\%, we find that a higher $m$ leads to better fusion of the visual knowledge in V-expert (see Section 4.4). Besides, choosing an appropriate $\mathcal{T}$ is also of importance for the adaptation process. Specifically, for V-expert we adopt a corpus with a stronger connection to visual concepts, e.g., image captions, while we adopt a corpus with richer language knowledge for T-expert. Therefore, we can separate the adaptation process for the V-expert and T-expert and find an optimal setting for them, respectively. Note that we freeze the original parameters of PLMs, which enables us to parallel the training of V- and T-expert. Besides, this significantly saves the memory footprint during adaptation, leading to great efficiency for the adaptation.

\begin{table*}[]
    \centering
        \begin{tabular}{cc}
            \toprule
            Prompts & Labels \\
            \midrule
            Q: What is the color of \texttt{[ITEM]}? A: It is \texttt{[MASK]}. &  \\
           Q: What is the colour of \texttt{[ITEM]}? A: It is \texttt{[MASK]}. & \\
           What is the color of \texttt{[ITEM]}? It is \texttt{[MASK]}. & \\
            What is the colour of \texttt{[ITEM]}? \texttt{[MASK]}. & \{blue, white, red, yellow, \\
            The color of \texttt{[ITEM]} is \texttt{[MASK]}. &  black, green, purple, brown, \\
            The usual color of \texttt{[ITEM]} is \texttt{[MASK]}. & pink, grey, orange \} \\
           \texttt{[ITEM]} usually has the color of \texttt{[MASK]}. & \\
            What is the usual color of \texttt{[ITEM]}? \texttt{[MASK]}. & \\
            What is the typical color of \texttt{[ITEM]}? \texttt{[MASK]}. & \\
            \bottomrule
        \end{tabular}
   \caption{Zero-shot reasoning prompt sample. For each input sentence, we replace the \texttt{[ITEM]} token with the input object and predict the color from the \texttt{[MASK]} token.}
   \label{tab: zsl-prompt}
\end{table*}

\paragraph{Fine-tuning} For downstream tasks on a single text, e.g., sentiment analysis, we directly input the visual features related to the input into X-adapters for knowledge fusion. For downstream tasks on multiple text, e.g., textual entailment, we first find the relevant visual features for them separately. Then, we stack them with a \texttt{TOKEN\_TYPE\_ID} to indicate which sentence the visual features belong to, as done in \cite{BERT}. 

\paragraph{Zero-shot reasoning}
For zero-shot reasoning tasks, we first construct prompts with a \texttt{[MASK]} token, as shown in \Cref{tab: zsl-prompt}. Then, we take the logits of the candidate classes at the \texttt{[MASK]} position as the predictions.

\section{Experiments}

\subsection{Setup}

\paragraph{Training data \& Benchmarks}
For V-expert, we adopt the image captions from MS COCO \cite{lin2014microsoft} as the training corpus, while T-expert is trained on Wiki103 \cite{merity2017pointer}.
To validate the effectiveness of our methods, we investigate two tasks: zero-shot object-color reasoning and natural language understanding (NLU). 
For the former, we adopt two benchmarks, \texttt{MemoryColor} \cite{norlund2021transferring} and \texttt{ColorTerms} \cite{bruni2012distributional}, which evaluate language models' abilities on reasoning the color of common objects. For the latter, we conduct experiments over seven benchmarks from GLUE \cite{wang2018glue}. We now detail these data and benchmarks.

\textbf{MS COCO} is a dataset including images and captions, proposed by \cite{lin2014microsoft}. We collect all the captions in it as the training corpus for V-expert. There are about 7M tokens and 0.6M sentences. It is released under a Creative Commons Attribution 4.0 License.

\textbf{Wiki103} is a featured subset of English Wikipedia proposed by \cite{merity2017pointer}. It includes 111M tokens and 4.2M sentences. We use Wiki103 as the training corpus for T-expert. Wiki103 is released under a Creative Commons Attribution-ShareAlike License.

\textbf{GLUE Benchmark} is a widely-used collection of benchmarks to evaluate models' capbilities on NLU tasks. GLUE consists of RTE \cite{bar2005definition}, MRPC \cite{dolan-brockett-2005-automatically}, STSB \cite{cer2017semeval}, CoLA \cite{warstadt-etal-2019-neural}, SST2 \cite{socher-etal-2013-recursive}, QNLI \cite{rajpurkar-etal-2016-squad}, QQP \cite{qqp} and MNLI \cite{williams-etal-2018-broad}. The individual datasets are released under different permissive licenses.

\textbf{MemoryColor} is a benchmark introduced by \cite{norlund2021transferring}. \texttt{MemoryColor} includes color information of 109 common objects, e.g., the grass. We adopt all the 109 samples to evaluate our method's understanding on visual knowledge. There are in total eleven colors as the candidate classes. This benchmark is released under a Creative Commons Attribution 4.0 International License.

\textbf{ColorTerms} is a benchmark introduced by \cite{bruni2012distributional}. \texttt{ColorTerms} includes color information of 53 common objects, and it has the same eleven candidate colors as \texttt{MemoryColor}. We leverage all the 53 samples to evaluate our method. This benchmark is released under a Creative Commons Attribution-NonCommercial-ShareAlike 3.0 International License.


\paragraph{Baselines} 
We adopt two masked language models, BERT \cite{BERT} and RoBERTa \cite{liu2019roberta}, as our baselines on both tasks. We insert our proposed X-adapter module into the two baseline models to validate the effectiveness of fusing external visual information. Besides, for zero-shot visual-related reasoning tasks, we compare our method with Voken \cite{vokenization}, and VaLM \cite{wang2022visually}. For NLU tasks, we compare with XDBert \cite{xdbert}.

\paragraph{Implementation Details}
For baseline BERT and RoBERTa models, we use the weight checkpoints released from HuggingFace \cite{wolf-etal-2020-transformers}. For X-adapters, the hidden dimension is set to 512, the intermediate hidden dimension in FFN is set to 2048 and the head number is set to 8 for the cross attention. 
We compare a single layer transformer encoder in BERT \cite{BERT} with X-adapter. As shown in \Cref{tab:arch-comparison}, one layer of X-adapter (T- or V-expert) has only 60\% of number of parameters in one layer of transformer in BERT. Taking the word embedding into consideration, inserting one (resp. two) layer of X-adapter into base-sized BERT only increase 3.7\% (resp. 7.5\%) parameters of the whole PLM. This parameter-efficiency leads to fast training during adaptation, since only the inserted parameters are tuned.  
We insert one layer of V-expert for color reasoning tasks and two layers of T-expert for NLU tasks.
As for training setup, we adopt Adam \cite{adam} as the optimizer with an initial learning rate of 1e-4. We train the V-expert and T-expert for three epochs on COCO caption and one epoch on Wiki103, respectively. The batch size is set to 256 and 96 for V- and T-expert, respectively.
For downstream zero-shot reasoning tasks, we adopt 9 prompts templates, same as VaLM \cite{wang2022visually}, as shown in \Cref{tab: zsl-prompt}. For downstream NLU tasks, we finetune the model for three epochs for all tasks. For BERT baselines, we follow the fine-tuning settings in XDBert \cite{xdbert}, as shown in \Cref{tab: xdbert-lr}. For RoBERTa, we use a consistent learning rate 2e-5 for all the datasets, since we find 1e-4 is not stable for RTE, MRPC and STS-B during training. It takes about 15 minutes and 45 minutes to train an epoch for V- and T-expert, respectively.
All the experiments are conducted on four Nvidia Tesla A100 GPUs with 80GB GPU memory.
\begin{table}[tb!]
    \centering
    \small
        \begin{tabular}{ccc}
            \toprule
            & $\mathrm{BERT}_{base}$ & X-adapter \\
            \midrule
            Hidden\_dim & 768 & 512\\
            Attn\_head & 12 & 8\\
            FFN\_dim & 3072 & 2048\\
            Total params & 7.1M & 4.2M \\
            \bottomrule
        \end{tabular}
   \caption{Detailed comparison on model architecture (\textbf{single} transformer layer).}
   \label{tab:arch-comparison}
\end{table}
\begin{table}[tb!]
    \centering
    \small
    {
        \begin{tabular}{ccc}
            \toprule
            Datasets & Base-sized & Large-sized \\
            \midrule
            RTE, MRPC, STSB & 1e-4 & 5e-5  \\
        Others & 2e-5 & 1e-5 \\
            \bottomrule
        \end{tabular}
    }
   \caption{Fine-tuning setting for BERT models.}
   \label{tab: xdbert-lr}
\end{table}

\subsection{Zero-shot object color reasoning}
\begin{table}[!tbp]
        \centering
        \small
            {
            \begin{tabular}{cccc}
                \toprule
                 Model & MC & CT & AVG  \\
                \midrule
                 $\mathrm{BERT}_{base}$ \cite{BERT} & 29.56 & 28.84  & 29.20 \\
                 $\mathrm{RoBERTa}_{base}$ \cite{liu2019roberta} & 34.05 & 30.98  & 32.52 \\
                 $\mathrm{Voken}\mathrm{(BERT}_{base})$ \cite{vokenization} & 14.27 & 19.01 & 16.64\\
                 $\mathrm{VaLM \text{-}4}$ \cite{wang2022visually}  & 53.99 & 52.66  & 53.33  \\
                 $\mathrm{VaLM \text{-}8}$ \cite{wang2022visually} & 58.64 & 50.19  & 54.47  \\
                 \midrule
                 $\mathrm{X\text{-}adapter(RoBERTa}_{base})$  & 59.63 & 53.85 & 56.74 \\
                 $\mathrm{X\text{-}adapter(BERT}_{base})$  & \textbf{64.11} & \textbf{60.04} & \textbf{62.08} \\
                 \midrule 
                 \midrule
                 $\mathrm{BERT}_{large}$ \cite{BERT} & 35.67 & 35.68  & 35.68 \\
                 \midrule
                 $\mathrm{X\text{-}adapter(BERT}_{large})$  & \textbf{66.56} & \textbf{63.25} & \textbf{64.90} \\
                \bottomrule
            \end{tabular}
            }
       \caption{Accuracy on zero-shot object color reasoning tasks. MC, CT and AVG denote \texttt{MemoryColor}, \texttt{ColorTerms} and average accuracy, respectively.}
       \label{tab: zsl-color-reasoning}
    \end{table}
\begin{table*}[!tbp]
    \centering
    \small
    {
    \centering
        \begin{tabular}{cccccccccc}
            \toprule
             Model & RTE & MRPC	& STS-B & CoLA	& SST-2 & QNLI & QQP & MNLI & AVG\\
            \midrule
             $\mathrm{BERT}_{base}$ \cite{BERT} &   67.07 & 87.47 &  89.19 & 56.50 & 92.29 & 91.13 & 89.51 & 84.47 &  82.20 \\
             $\mathrm{XDBert(BERT}_{base})$ \cite{xdbert}  & 69.31 & 88.02 & \textbf{89.32} & \textbf{57.55} & \textbf{92.78} & 
             \textbf{91.52} & \textbf{89.57} & \textbf{84.75} & \textbf{82.85}\\
             $\mathrm{X\text{-}adapter(BERT}_{base})$  & \textbf{70.90} & \textbf{88.15} & 88.91 & 56.66 & 92.75 & 91.47 & 89.35 & 84.38 & \textbf{82.82} \\
             \midrule
             $\mathrm{RoBERTa}_{base}$ \cite{liu2019roberta} & 69.43 & 89.02 & 89.63 & 57.28 & 94.15 & 92.10 & 89.74 & 87.57 & 83.62\\
             $\mathrm{X\text{-}adapter(RoBERTa}_{base})$  & \textbf{71.12} & \textbf{90.03}  & \textbf{89.93} & \textbf{58.50} & \textbf{94.55} & \textbf{92.63} & \textbf{89.90} &  \textbf{87.85} & \textbf{84.31} \\
            \midrule
              $\mathrm{BERT}_{large}$ \cite{BERT} & 70.47 & 87.94 & \textbf{89.75} & 57.33 & 93.14 & 91.64 & 89.63 & 86.22 & 83.27 \\
             $\mathrm{X\text{-}adapter(BERT}_{large})$  & \textbf{73.65} & \textbf{88.62} & 89.69 & \textbf{58.11} & \textbf{93.51} & \textbf{92.03} & \textbf{89.69} & 
                \textbf{86.32} & \textbf{83.95} \\
            \bottomrule
        \end{tabular}
    }
   \caption{The performance on GLUE. We report the average results on 5 runs and the macro average value over all the benchmarks.} 
   \label{tab: glue}
\end{table*}
We activate V-expert in X-adapters for zero-shot object color reasoning tasks. 
The results are shown in \Cref{tab: zsl-color-reasoning}. As we can see, the baseline BERT and RoBERTa model of base size (12L/768H) can only achieve approximately 30\% average accuracy on these two benchmarks, with 11 colors as the candidate classes. This poor performance is due to the lack of visual information during pre-training. The previous distillation-based method, Voken \cite{vokenization}, performs even worse than the baseline model, indicating that distillation from the text encoder of VLMs is not sufficient for tasks requiring vital visual information, e.g., reasoning the color of common objects. However, our method, X-adapter, can significantly alleviate this issue by injecting the relevant  image features into the language models, with 32.88\% and 24.22\% improvements on BERT and RoBERTa base models, respectively. Besides, compared with VaLM \cite{wang2022visually} that fuses image features during pre-training, our method (based on either BERT or RoBERTa of base size) still outperforms it with a large margin, with much less computational cost. Further, we also conduct experiments on larger baseline models, e.g., a 24-layer large BERT model (24L/1024H). We can still observe significant improvements on both benchmarks, suggesting that our method is effective across different model architectures. 

\subsection{Natural Language Understanding}
For NLU tasks, we activate T-expert in X-adapters. The results are shown in \Cref{tab: glue}. As we can see, X-adapters can improve the baseline base models on most of all the downstream tasks, with 0.62\% and 0.69\% improvements on the average accuracy for BERT and RoBERTa, respectively. Besides, for the 24-layer large BERT model, X-adapters surpass the baseline on all the downstream tasks except STS-B, achieving a 0.68\% improvement on average accuracy. Compared with the previous method XDBert \cite{xdbert}, our method achieves comparable performance on average accuracy with much better efficiency (tuning two models vs. tuning two layers). All the results demonstrate the effectiveness and efficiency of our method on NLU tasks.
\subsection{Ablation Studies}
In this section, we conduct ablation studies to further illustrate the effectiveness of our method. Due to the computational budget, all the experiments are done based on base-sized BERT.

\begin{table*}[tb!]
    \centering
    \small
    {
    \centering
        \begin{tabular}{cccccccccc|ccc}
            \toprule
             Model & RTE & MRPC	& STS-B & CoLA	& SST-2 & QNLI & QQP & MNLI & $\mathrm{AVG}_G$ & MC & CT & $\mathrm{AVG}_C$ \\
            \midrule
             V-expert-Wiki & 65.74	 & 87.98 &  88.83 & 56.51 & 92.59 & 91.16 & 89.19 & 84.34 &  82.04 & 32.11  & 33.33 & 32.72 \\
             V-expert-COCO & 67.12 & 86.63 & 88.69 & 56.07 & 92.62 & 90.91 & 89.16 & 83.96 & 81.90 & \textbf{64.11} & \textbf{60.04} & \textbf{62.08} \\
             T-expert-Wiki & \textbf{70.90} & \textbf{88.15} & \textbf{88.91} & \textbf{56.66} & 92.75 & 
             \textbf{91.47} & \textbf{89.35} & \textbf{84.38} & \textbf{82.82} & 37.00 & 32.48 & 
 34.74  \\
             T-expert-COCO & 65.95 & 86.99  &  88.57 & 55.31 & \textbf{92.82} & 90.42 & 89.11 & 83.81 & 81.62 & 37.00 & 38.24 & 37.62 \\
            \bottomrule
        \end{tabular}
    }
   \caption{Ablation study for experts with different training corpus. The baseline model is $\mathrm{BERT}_{base}$. $\mathrm{AVG}_G$, MC, CT, and $\mathrm{AVG}_C$ denote the average accuracy on GLUE, \texttt{MemoryColor} , \texttt{ColorTerms} and the average accuracy on color reasoning tasks, respectively.} 
   \label{tab:corpus-and-tasks}
\end{table*}

\paragraph{Experts and Training Corpus} First, we study the training corpus' effect on X-adapters and the best assignment for different experts in X-adapters and tasks. As shown in \Cref{tab:corpus-and-tasks}, for both V-expert and T-expert, training on COCO captions performs better on color reasoning tasks, while training on Wiki103 performs better on NLU tasks. We conjecture that this is because the COCO captions include more visual-related information, e.g., the appearance of objects, while Wiki103 contains more diverse and complicated semantic knowledge. Further, we can also observe that V-expert trained on COCO captions and T-expert trained on Wiki103 attain optimal performance on color reasoning and NLU tasks, respectively. Therefore, without further notice, we activate V-expert for color reasoning tasks and T-expert for NLU tasks in this paper.   

\paragraph{Effectiveness of CLIP features}
To verify the effectiveness of the input image and text features from CLIP, we conduct experiments on the following settings: 1. fine-tuning BERT baseline model with the same steps and corpus as we train the X-adapters (BERT-FT); 2. inserting X-adapters into BERT baseline model and fine-tuning without the CLIP features (BERT-FT-Param); 3. Inserting X-adapters into BERT baseline model and training with CLIP features, but no CLIP features during inference (w/o. CLIP); 4. Inserting X-adapters into BERT baseline model and training with CLIP features, and input random noise during inference instead of CLIP features (w. noise). Note that for setting 2 and 3, \texttt{cross-attention} in X-adapter will degenerate to \texttt{self-attention} since we do not impose any external features during inference. 

The results are shown in \Cref{tab:ablation-clip-image-feat}. First, the results of setting 1 (\nth{2} row) show that tuning all the parameters of the PLMs on small datasets, e.g., COCO captions and Wiki103, disturbs the knowledge learned on massive corpus and leads to performance degradation on both tasks. Then, setting 2 (\nth{3} row) outperforms the setting 1 slightly, indicating that learning in an adapter fashion (freezing the pre-trained model and tuning the added parameters) can integrate new knowledge into the PLMs. Further, X-adapters inject the visual knowledge from CLIP's text encoder and image encoder via T-expert and V-expert, respectively. With the external knowledge, X-adapters improve the baseline's performance by a large margin (\nth{6} row), especially for color reasoning tasks, demonstrating that the multi-modal information from CLIP is of importance. Moreover, the results of setting 3 (\nth{4} row) and setting 4 (\nth{5} row) imply that only adopting CLIP features during training, which is a variant of distillation \cite{xdbert}, is insufficient for X-adapters. Missing correct visual features during inference can lead to marginal improvement or even performance degradation.    
\begin{table*}[tb!]
    \centering
    \small
    {
    \centering
        \begin{tabular}{cccccccccc|ccc}
            \toprule
            Model & RTE & MRPC	& STS-B & CoLA	& SST-2 & QNLI & QQP & MNLI & $\mathrm{AVG}_G$ & MC & CT & $\mathrm{AVG}_C$ \\
            \midrule
            BERT & 67.07 & 87.47 & \textbf{89.19} & 56.50 & 92.29 & 91.13 & \textbf{89.51} & \textbf{84.47} & 82.20 & 29.56  & 28.84 & 29.20 \\
            BERT\text{-}FT & 66.88 & 87.06 & 89.15 & 53.87 & 91.93 & 90.80 & 89.22 & 84.31 & 81.65 & 24.26 & 29.27 &  26.77 \\
            BERT\text{-}FT-Param & 67.51 & 87.30 & 88.93 & 56.26 & 92.28 & 91.27 & 89.50 & 83.97 & 82.13 & 29.87 & 30.56 & 30.22 \\
            X\text{-}Adapter (w/o. CLIP) & 69.57 & 87.24 & 89.27 & 55.57 & 92.27  & 91.16 & 89.46 & 84.16 & 82.34 & 16.21  & 14.74	& 15.48\\
            X\text{-}Adapter (w. Noise) & 67.67 & 86.65 & 89.11 & 56.16 & 92.61 & 91.29 &  89.43 & 84.18 & 82.14 &  22.94 & 19.87 & 21.41  \\
            X\text{-}Adapter & \textbf{70.90} & \textbf{88.15} & 88.91 & \textbf{56.66} & \textbf{92.75} & \textbf{91.47} & 89.35 & 84.38 & \textbf{82.82} & \textbf{64.11} & \textbf{60.04} & \textbf{62.08} \\
            \bottomrule
        \end{tabular}
    }
   \caption{Ablation study on the effectiveness of CLIP features. The baseline model is $\mathrm{BERT}_{base}$. $\mathrm{AVG}_G$, MC, CT, and $\mathrm{AVG}_C$ denote the average accuracy on GLUE, \texttt{MemoryColor} , \texttt{ColorTerms} and the average accuracy on color reasoning tasks, respectively.} 
   \label{tab:ablation-clip-image-feat}
\end{table*}

\paragraph{Mask ratio} We conduct an ablation study on the mask ratio of MLM, as shown in \Cref{fig: mlm-ratio} and \Cref{tab: mlm-t-expert-full}. For V-expert, different from the widely used mask ratio 15\%, larger mask ratio leads to better fusion of visual knowledge. In our case, 45\% mask ratio achieves the best performance on both benchmarks. This implies that the visual features imposed by V-expert are beneficial such that higher mask ratios are bearable. However, for T-expert, we find that the widely used 15\% mask ratio is the optimal. 

\begin{table*}[]
    \centering
    \small
    {
    \centering
        \begin{tabular}{cccccccccc|ccc}
            \toprule
             Mask Ratio & RTE & MRPC	& STS-B & CoLA	& SST-2 & QNLI & QQP & MNLI & $\mathrm{AVG}_G$ & MC & CT & $\mathrm{AVG}_C$\\
            \midrule
             0.05 & 67.15 & 87.05 & 88.90 & 55.47  & 92.73 & 91.32  & 89.06 & 84.21 & 81.99 & 43.53 & 43.80 & 43.67 \\
             0.15 & \textbf{70.90} & \textbf{88.15} & \textbf{88.91} & 56.66 & \textbf{92.75} & \textbf{91.47} & \textbf{89.35} & \textbf{84.38} & \textbf{82.82} & 62.79 & 55.55 & 59.17 \\
             0.25 & 69.45 & 86.90 & 88.88 & \textbf{57.46} & 92.66 & 91.23 & 89.18 & 84.23 & 82.50 &  63.51 & 56.83 & 60.17  \\
             0.35 & 68.47 & 86.95  & 88.67 & 56.50 & 92.71 & 91.21 & 89.34 & 84.28 & 82.27 & 63.51 & 58.76 & 61.14 \\
             0.45 & 66.73 & 87.28 & 88.86 & 55.40 & 92.51 & 91.32 &  89.28 & 84.29 & 81.97  & \textbf{64.11} & 60.04 & \textbf{62.08} \\
             0.55 & 67.38 & 86.53 & 88.58 & 55.42 & 92.62 & 91.37 & \textbf{89.35} & 84.32 & 81.96 & 63.91 & \textbf{60.09} & 62.00 \\
             0.65 &  66.71 & 86.93 & 88.71 & 55.37	 & 92.68 & 91.21  & 89.33  & 84.23 & 81.90  & 63.92 & 59.70 & 61.81  \\
            \bottomrule
        \end{tabular}
    }
   \caption{Ablation study on the mask ratio. The baseline model is $\mathrm{BERT}_{base}$. $\mathrm{AVG}_G$, MC, CT, and $\mathrm{AVG}_C$ denote the average accuracy on GLUE, \texttt{MemoryColor} , \texttt{ColorTerms} and the average accuracy on color reasoning tasks, respectively.} 
   \label{tab: mlm-t-expert-full}
\end{table*}

\begin{figure}[tb!]
    \centering
    \hspace{-2.5cm}
    \subfloat[]{ 

\definecolor{myblue}{RGB}{29,114,221}    
\definecolor{myyellow}{RGB}{255,255,191} 
\definecolor{myorange}{RGB}{244,106,18}  
\definecolor{mygray}{RGB}{102,102,102}   
\definecolor{mypink}{RGB}{252,228,215}   

\definecolor{CUpurple}{RGB}{117,15,109}
\definecolor{CUlpurple}{RGB}{165,133,182}
\definecolor{CUgold}{RGB}{221,163,0}
\definecolor{CUribbon}{RGB}{244,223,176}
\definecolor{CUblack}{RGB}{34,24,21}
\definecolor{PKUred}{RGB}{126,24,28}

\definecolor{Green-d}{RGB}{81,157,62}
\definecolor{Green-l}{HTML}{d2e7ce}
\definecolor{Blue-d}{HTML}{3a76af}
\definecolor{Blue-l}{HTML}{d2dff0}
\definecolor{Orange-d}{HTML}{f08536}
\definecolor{Orange-l}{HTML}{fbe6d1}

\pgfplotsset{
    width =4.4cm,
    height=3.3cm,
}


\begin{tikzpicture}[scale=1]
    \begin{axis}[
    xmin=0.5, xmax=6.5,
    ymin=81.8, ymax=83.0,
    xtick={1,2,3,4,5,6},
    xticklabels={0.05,0.15,0.25,0.35,0.45,0.55}, 
    x tick label style = {font=\small, rotate=45},
    ytick = {82.0, 82.5, 83.0},
    yticklabels = {82.0, 82.5, 83.0},
    y tick label style = {font=\small},
    ylabel={Accuracy},
    y label style={at={(-0.23,0.5)}, font=\small},
    legend style={
        draw=none,
        at={(0.01,1.0)},
        anchor=south west,
        legend columns=2,
        font=\small
    }
    ]
        \addplot[red, style={mark=o, mark size=1.0pt, line width=1.0pt, draw=Green-d}]   table {glue.dat};
        \legend{$\mathrm{AVG}_G$}
    \end{axis}
\end{tikzpicture}
 \label{fig: mlm-t-expert} }
    \hspace{-2.5cm}
    \subfloat[]{ 

\definecolor{myblue}{RGB}{29,114,221}    
\definecolor{myyellow}{RGB}{255,255,191} 
\definecolor{myorange}{RGB}{244,106,18}  
\definecolor{mygray}{RGB}{102,102,102}   
\definecolor{mypink}{RGB}{252,228,215}   

\definecolor{CUpurple}{RGB}{117,15,109}
\definecolor{CUlpurple}{RGB}{165,133,182}
\definecolor{CUgold}{RGB}{221,163,0}
\definecolor{CUribbon}{RGB}{244,223,176}
\definecolor{CUblack}{RGB}{34,24,21}
\definecolor{PKUred}{RGB}{126,24,28}

\definecolor{Green-d}{RGB}{81,157,62}
\definecolor{Green-l}{HTML}{d2e7ce}
\definecolor{Blue-d}{HTML}{3a76af}
\definecolor{Blue-l}{HTML}{d2dff0}
\definecolor{Orange-d}{HTML}{f08536}
\definecolor{Orange-l}{HTML}{fbe6d1}

\pgfplotsset{
    width =4.4cm,
    height=3.3cm,
}


\begin{tikzpicture}[scale=1]
    \begin{axis}[
    xmin=0.5, xmax=6.5,
    ymin=58, ymax=63,
    xtick={1,2,3,4,5,6},
    xticklabels={0.15,0.25,0.35,0.45,0.55,0.65}, 
    x tick label style = {font=\small, rotate=45},
    ytick = {58, 60, 62},
    yticklabels = {58, 60, 62},
    y tick label style = {font=\small},
    ylabel={Accuracy},
    y label style={at={(-0.16,0.5)}, font=\small},
    legend style={
        draw=none,
        at={(0.01,1.0)},
        anchor=south west,
        legend columns=2,
        font=\small
    }
    ]   
        \addplot[red, style={mark=o, mark size=1.0pt, line width=1.0pt, draw=Green-d}]   table {avg.dat};
        \legend{$\mathrm{AVG}_C$}
    \end{axis}
\end{tikzpicture}
 \label{fig: mlm-v-expert} }
    \caption{Abalation study on the mask ratio. (a) Performance for T-expert; (b) Performance for V-expert.}
    \label{fig: mlm-ratio}
\end{figure}
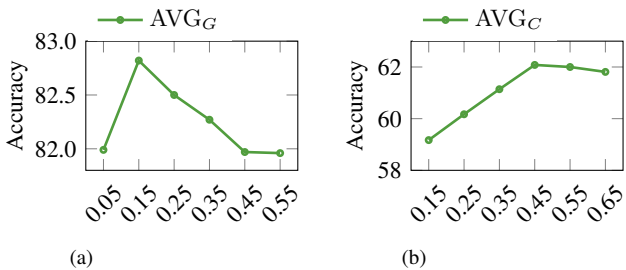

\paragraph{Insertion positions}
We also conduct an ablation study on the positions where we insert X-adapters. For simplicity, we insert one layer of X-adapter (V- and T-expert) into different positions in the transformer encoder (before the \nth{3}, \nth{6}, \nth{9}, and \nth{12} transformer layers). As shown in \Cref{tab:insertion-position}, insertion before the last transformer layer (\nth{12} layer) achieves the best performance for both V- and T-expert. As the position goes shallow, the performance drops. This finding is consistent with \cite{rogers2020primer} that features at deeper layers contain richer semantic knowledge and are thus easier to fuse with features from CLIP.
\begin{table*}[]
    \centering
    \resizebox{.88\linewidth}{!}
    {
    \centering
        \begin{tabular}{cccccccccc|ccc}
            \toprule
             Position & RTE & MRPC	& STS-B & CoLA	& SST-2 & QNLI & QQP & MNLI & $\mathrm{AVG}_G$ & MC & CT & $\mathrm{AVG}_C$ \\			
            \midrule
             3 & 64.19 & 84.69 & 87.81  &56.40  & 92.55 & 91.01 & 89.46& \textbf{84.40} &  81.31 & 46.28 & 44.66 & 45.47\\
             6 & 66.50 & 86.63 & 88.83  & 56.25 & 92.62 & \textbf{91.35} & 89.26 & 84.32 & 81.97 & 46.18 & 45.94 & 46.06 \\
             9 & 66.83	& \textbf{87.41} & 88.76 & 56.52 & \textbf{92.82} & 91.15	& 89.27 & 84.20	& 82.12 & 50.87 & 48.08 & 49.48 \\				
             12 & \textbf{69.64} & 86.77	 & \textbf{89.29} & \textbf{56.64} & 92.41	 & 91.10 & \textbf{89.54} & \textbf{84.40} & \textbf{82.47} & \textbf{64.11} & \textbf{60.04} & \textbf{62.08} \\
            \bottomrule
        \end{tabular}
    }
   \caption{Ablation study on the adapter position. The baseline model is $\mathrm{BERT}_{base}$. $\mathrm{AVG}_G$, MC, CT, and $\mathrm{AVG}_C$ denote the average accuracy on GLUE, \texttt{MemoryColor} , \texttt{ColorTerms} and the average accuracy on color reasoning tasks, respectively.} 
   \label{tab:insertion-position}
\end{table*}

\paragraph{Number of layers} We conduct the ablation study on the number of layers of X-adapters, as shown in \Cref{tab: layer-num-t-expert-full}. We append some X-adapter layers before the last several transformer layers. For T-expert, two X-adapter layers achieve the best performance. However, increasing the number of layers does not lead to further improvement. For V-expert, appending only one layer of X-adapter before the last transformer layer achieves the best performance.  We conjecture that this is due to the large domain gap between the image and the text features. Involving too many image features harms the language knowledge learned in PLMs. 
\begin{table*}[]
    \centering
    \resizebox{.9\linewidth}{!}
    {
    \centering
        \begin{tabular}{cccccccccc|ccc}
            \toprule
             \# of layers & RTE & MRPC & STS-B & CoLA & SST-2 & QNLI & QQP & MNLI & $\mathrm{AVG}_G$ & MC & CT & $\mathrm{AVG}_C$ \\			
            \midrule
             1 & 70.64 & 86.97	 & 88.79	&56.14 & 92.41	 & 91.10 & 89.54 & 84.20 & 82.47 & \textbf{64.11} & \textbf{60.04} & \textbf{62.08}  \\
             2 & \textbf{70.90} & \textbf{88.15} & \textbf{88.91} & 56.66 & \textbf{92.75} & \textbf{91.47} & 89.35 & \textbf{84.38} & \textbf{82.82} & 63.79 & 56.59  & 60.19  \\
             3 & 68.25 & 87.57& 88.72 & \textbf{57.83} & 92.68 & 91.35 & 89.36 & 84.20  &  82.50 & 61.75  & 55.09  & 58.42 \\
            \bottomrule
        \end{tabular}
    }
   \caption{Ablation study on number of layers of X-adapters. The baseline model is $\mathrm{BERT}_{base}$. $\mathrm{AVG}_G$, MC, CT, and $\mathrm{AVG}_C$ denote the average accuracy on GLUE, \texttt{MemoryColor} , \texttt{ColorTerms} and the average accuracy on color reasoning tasks, respectively.} 
   \label{tab: layer-num-t-expert-full}
\end{table*}

\paragraph{V-expert: Image bank size}
The size of the image banks is essential for the quality of image retrieval. Since we construct our image bank with the training and validation set of COCO \cite{lin2014microsoft}, and the whole Visual Genome \cite{krishna2017visual} dataset, we investigate four combinations with different sizes of the image bank, as shown in \Cref{tab:zsl-img-sets}. As the size of the image bank increases, the quality and diversity of the retrieved images improve, leading to an improvement of the zero-shot reasoning accuracy.

\begin{table*}[!htbp]
    \centering
    {
    \centering
        \begin{tabular}{c|ccc|c|ccc}
            \toprule
             \multirow{2}{*}{Model} & \multicolumn{3}{c|}{Image Sets} & \multirow{2}{*}{Img \#} & \multirow{2}{*}{\texttt{MemoryColor}} &
             \multirow{2}{*}{\texttt{ColorTerms}} &
             \multirow{2}{*}{AVG}\\
             & $\mathrm{COCO}_{V}$ & $\mathrm{COCO}_{T}$ & VG & &  \\
              \midrule
              \multirow{4}{*}{V\text{-}expert} & \checkmark &   & & 40K & 59.61 & 53.81  & 56.71 \\
              & \checkmark &   &\checkmark & 97K & 60.53 & 55.95 & 58.24 \\
             & \checkmark & \checkmark &  & 120K & 64.40 & 58.72 & 61.56 \\
             & \checkmark & \checkmark & \checkmark & 170K & 64.11 & 60.04 & 62.08 \\
            \bottomrule
        \end{tabular}
    }
   \caption{Ablation study on the size of the image bank. $\mathrm{COCO}_T$, $\mathrm{COCO}_V$ and VG denote the training set of COCO, the validation set of COCO and Visual Genome (excluding the images in COCO), respectively.}
   \label{tab:zsl-img-sets}
\end{table*}

\begin{figure}[tb!]
    \centering
    \input{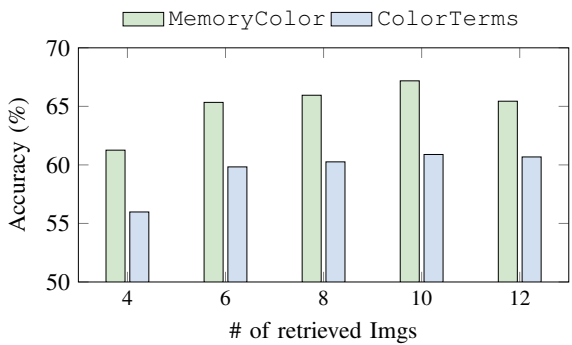}
    \caption{Ablation study on the number of images retrieved for the input text.}
    \label{fig:top-k}
\end{figure}

\paragraph{V-expert: Number of retrieved images} The number of retrieved images for each input text sequence is another important factor impacting the visual knowledge injected into the language models. We conduct experiments on the number of retrieved images, as shown in \Cref{fig:top-k}. As we increase $K$ from 4 to 10, the accuracy on the two benchmarks improves. When $K > 10$, the accuracy on the two datasets starts to decrease, which indicates that too few retrieved images can not bring enough visual knowledge, while too many may lead to redundancy and performance degradation. $K=10$ is the best choice in our scenario. 

\subsection{Analysis}
\paragraph{The robustness of image retrieval in V-expert.} 
Since in real scenario there may be noise in the image bank,
we test our method's robustness by introducing low-quality images in the image bank. Specifically, we augment the images in the original image bank by random crop, color jittering, random perspective transform, and Gaussian blur. We set the number of the noisy images to be the same as that of the clean images, and combine the noisy and clean images as a new noisy image bank. As shown in \Cref{tab:robustness}, with many noisy images, V-expert with image retrieval still achieves good performance on color reasoning tasks (accuracy loss less than 1\%), validating the robustness of our proposed method.  
\begin{table}[!tbp]
    \centering
    \small
    %
    {
        \begin{tabular}{cccc}
            \toprule
             Image bank & MC & CT & AVG \\
              \midrule
             Clean & 64.11 & 60.04 & 62.08 \\
             Noisy & 63.71 & 59.19 & 61.45 \\
            \bottomrule
        \end{tabular}
    }
   \caption{Performance of V-expert ($\text{BERT}_{base}$) on noisy image bank. MC, CT and AVG denote \texttt{MemoryColor}, \texttt{ColorTerms} and average accuracy, respectively.}
   \label{tab:robustness}
\end{table}

\begin{figure}[tb!] 
    \centering
    \includegraphics[width=0.9\linewidth]{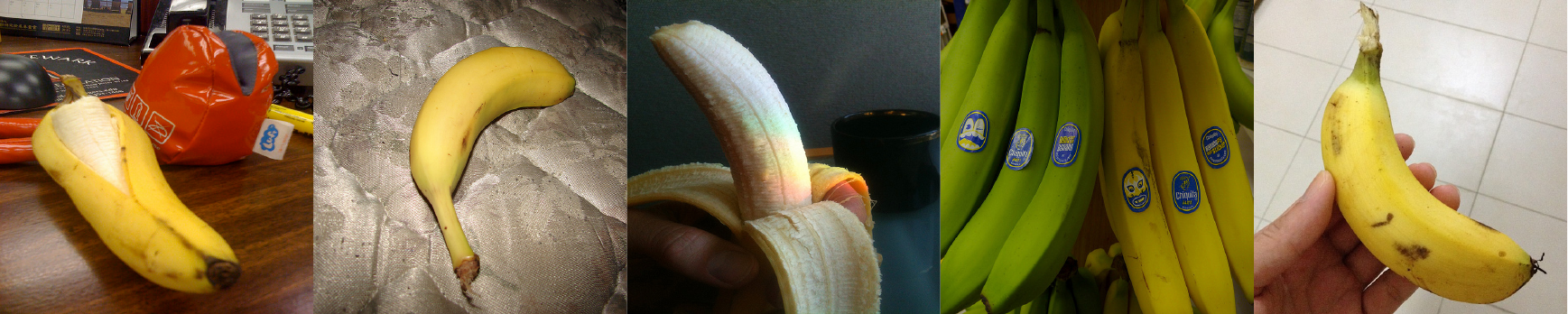}
    \caption{Top-5 relevant images retrieved for the prompt "What is the color of the banana? It is [MASK]."}
    \label{fig:retrieved-img}
\end{figure}

\begin{figure}[tb!]
    \centering
    \definecolor{mypink}{RGB}{255,171,164}     
\definecolor{mygray}{RGB}{240,240,240}     
\definecolor{lightblue}{RGB}{221,239,250}  
\definecolor{myred}{RGB}{228,46,36}        
\definecolor{mygreen}{RGB}{142,202,206}    
\definecolor{myyellow}{RGB}{255,253,181}   
\definecolor{mypurple}{RGB}{71,162,241}      
\definecolor{myorange}{RGB}{244,106,18}
\definecolor{mylight}{RGB}{191,191,255}
\definecolor{darkblue}{HTML}{226E9C}
\definecolor{crystal}{HTML}{39B185}
\definecolor{crybrown}{RGB}{160,127,40}
\definecolor{CUpurple}{RGB}{117,15,109}
\definecolor{CUlpurple}{RGB}{165,133,182}
\definecolor{CUgold}{RGB}{221,163,0}
\definecolor{CUribbon}{RGB}{244,223,176}
\definecolor{CUblack}{RGB}{34,24,21}
\definecolor{PKUred}{RGB}{126,24,28}

\definecolor{Green-d}{RGB}{81,157,62}
\definecolor{Green-l}{HTML}{d2e7ce}
\definecolor{Blue-d}{HTML}{3a76af}
\definecolor{Blue-l}{HTML}{d2dff0}
\definecolor{Orange-d}{HTML}{f08536}
\definecolor{Orange-l}{HTML}{fbe6d1}
\definecolor{Purple}{HTML}{e4d8e8}
\definecolor{Blue}{HTML}{d2dff0}
\definecolor{Yellow}{HTML}{fff3d2}
\definecolor{Green}{HTML}{d2e7ce}
\definecolor{Red}{HTML}{eec6c5}

\pgfplotsset{
    width =0.98\linewidth,
    height=0.52\linewidth
}
\begin{tikzpicture}[scale=.9]
    \begin{axis}[
        ybar,
        xmin=0.5, xmax=3.5,
        xlabel={Color},
        xlabel near ticks,
        xticklabels={blue, red, yellow},
        xtick={1, 2, 3},
        xtick align=inside,
        xticklabel style = {rotate=0}, xticklabel style = {font=\small},
        ymin=0, ymax=15, 
        ytick={3, 6, 9, 12, 15},
        yticklabels={3, 6, 9, 12, 15},
        bar width = 8pt,
        ylabel={Logits},
        ylabel near ticks,
        legend style={
            draw=none,
            at={(1.0,1.0)},
            anchor=north west,
            legend columns=1,
        }
        ]
        \addplot +[ybar, fill=Purple, draw=black, area legend]       table [x={Model},  y={Search}]     {blue-red.dat};
        \addplot +[      fill=Blue, draw=black, area legend]         table [x={Model},  y={Search}]     {blue.dat};
        \addplot +[      fill=Red, draw=black, area legend]         table [x={Model},  y={Search}]     {red.dat};
        \addplot +[      fill=Yellow, draw=black, area legend]         table [x={Model},  y={Search}]     {ret.dat};
        \addplot +[      fill=Green, draw=black, area legend]         table [x={Model},  y={Search}]     {base.dat};
        \legend{BR,B,R,Ret,Base}

        
    \end{axis}
\end{tikzpicture}
    \caption{Prediction logits for different input image features. BR, B, R, Ret and Base denote a mix of blue and red images, blue images, red images, retrieved images and BERT baseline, respectively. We show the logits for blue, red and yellow. The largest logit among these three colors is also the largest logit among all the eleven candidate colors for all the settings.}
    \label{fig:color-analysis}
\end{figure}
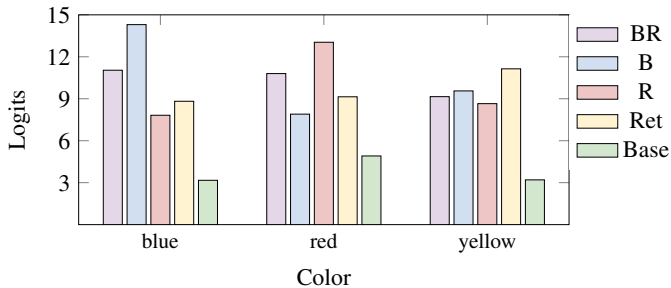

\paragraph{How well do X-adapters understand the visual concepts?}
We conduct additional analysis to further study how well the model understands the visual concepts and how it affects the predictions. We take the motivating example of questioning the color of the banana, for instance. For the prompt "What is the color of the banana? It is \texttt{[MASK]}.", red is the prediction with the highest probability of BERT baseline model. X-adapters will retrieve top-$K$ most relevant images, as shown in \Cref{fig:retrieved-img}, to the prompt and fuse the features from CLIP\_I into PLMs, leading to a correct prediction of the color yellow. Further, we insert images with different colors to verify whether X-adapter can understand these color concepts and change the final prediction. Three settings are considered: (1) all blue images, (2) all red images, and (3) a mix of blue and red images. The predictions are shown in \Cref{fig:color-analysis}. First, the vanilla BERT baseline model makes a wrong prediction (red) due to the loss of visual commonsense knowledge during pre-training, while X-adapter corrects the prediction (yellow) with retrieved relevant images. Then, when inserting all blue or red images, the logit of blue or red becomes the largest. In addition, when the input is a mix of blue and red images, we can find that the logits of blue and red are approximately the same and greater than other colors. This indicates that X-adapter does understand the visual concepts in the image features and can fully utilize them to improve the final predictions of LMs.    

\section{Conclusion}
In this paper, we propose a plug-and-play module, X-adapter, to inject the visual knowledge from pre-trained VLMs into PLMs. There are two sub-modules in the X-adapter, namely V-expert and T-expert, to integrate the features from VLM's image and text encoder, respectively. The X-adapters can fully exploit the competence of VLMs with an efficient adaptation process that only updates a few parameters. By activating different sub-modules, X-adapters allow us to flexibly fuse features in different modalities from VLMs for different downstream tasks. Extensive experimental results demonstrate that our method can significantly outperform the baseline language models on both object-color reasoning and NLU tasks.    

{
    \bibliographystyle{IEEEtran}
    \bibliography{ref/top,ref/nlp,ref/anthology,ref/custom}
}

\end{document}